\documentclass[a4paper,11pt,twocolumn,twoside]{article}
\usepackage{graphicx}
\usepackage{sepln}
\usepackage{fullname_esp}
\usepackage[spanish,es-nosectiondot, es-tabla, es-noindentfirst]{babel}
\usepackage[utf8]{inputenc}
\usepackage{amsmath}
\usepackage{booktabs}

\makeatletter
\newcommand{\@BIBLABEL}{\@emptybiblabel}
\newcommand{\@emptybiblabel}[1]{}
\makeatother
\usepackage{hyperref}

\DeclareMathOperator*{\argmax}{\arg\!\max}

\input epsf

\title{RETUYT en TASS 2017: Análisis de Sentimiento de Tweets en Español utilizando SVM y CNN}

\author {
Aiala Rosá \qquad Luis Chiruzzo \qquad Mathias Etcheverry \qquad Santiago Castro \\
Facultad de Ingeniería, \\
Universidad de la República \\
Montevideo, Uruguay \\
\{aialar, luischir, mathiase, sacastro\}@fing.edu.uy
}

\seplntranstitle{RETUYT in TASS 2017: Sentiment Analysis for Spanish Tweets using SVM and CNN}

\seplnclave{Análisis de Sentimiento, SVM, Redes Neuronales, Word Embeddings}

\seplnresumen{En este artículo se presentan clasificadores basados en SVM y Redes Neuronales Convolucionales (CNN) para la competencia TASS 2017 de clasificación de sentimientos de tweets. El clasificador con mejor desempeño en general utiliza una combinación de SVM y CNN. El uso de \textit{word embeddings} resultó particularmente importante para mejorar el desempeño de los clasificadores.}

\seplnkey{Sentiment Analysis, SVM, Neural Networks, Word Embeddings
}

\seplnabstract{This article presents classifiers based on SVM and Convolutional Neural Networks (CNN) for the TASS 2017 challenge on tweets sentiment analysis. The classifier with the best performance in general uses a combination of SVM and CNN. The use of word embeddings was particularly useful for improving the classifiers performance.}

\firstpageno{1}

\begin{document}

\label{firstpage}

\maketitle

\section{Introducción}

Dentro del área análisis de sentimiento, el análisis de tweets resulta especialmente interesante debido al gran volumen de información que se genera diariamente, la naturaleza subjetiva de la mayoría de los mensajes, y el fácil acceso a este material para su análisis y procesamiento. La existencia de tareas específicas vinculadas a este problema desde hace ya varios años evidencia el interés de la comunidad del Procesamiento de Lenguaje Natural en trabajar en este tema.

Desde el año 2012 se viene realizando el Taller de Análisis de Semántico de la SEPLN (TASS)\footnote{\url{http://www.sepln.org/workshops/tass}}, que está centrado en la clasificación de tweets. También en el marco de la SemEval\footnote{\url{https://aclweb.org/aclwiki/SemEval_Portal}}, desde el año 2013 se incluye una tarea sobre análisis de sentimiento en tweets, abordando el problema para el idioma inglés, y a partir de este año también para el árabe. Los resultados reportados en las últimas ediciones de estas tareas \cite{garcia2016overview,SemEval:2017:task4} no alcanzan aún niveles satisfactorios, principalmente para el español, por lo que el problema sigue resultando desafiante y se hace necesario seguir trabajando en él.

Los trabajos que obtuvieron los mejores resultados en la clasificación de tweets en español en la edición 2016 del TASS se basaron en la definición de atributos para entrenar varios clasificadores y luego combinarlos. En~\cite{Ceron2016JACERONG} se describe la construcción de un conjunto de clasificadores de tipo Regresión Logística que combinan diferentes atributos, entre los cuales se incluyen pertenencia a un léxico subjetivo, procesamiento de la negación, información morfo-sintáctica y atributos basados en n-gramas. En~\cite{Hurtado2016ELiRFUPVET} se entrena un conjunto de clasificadores de tipo SVM utilizando atributos morfo-sintácticos y atributos basados en n-gramas. Otros trabajos presentados incluyen el uso de \textit{word embeddings} \cite{Montejo-Raez2016,Quiros2016}, alcanzando resultados un poco más bajos. Por otro lado, la mayoría de los enfoques presentados en la tarea 4 de la SemEval 2017, aplicados para el idioma inglés, se basan en redes neuronales profundas.

En este artículo se describen los sistemas presentados al TASS 2017~\cite{tass2017} por el grupo RETUYT. Se abordó la problemática principalmente mediante el uso de \textit{word embeddings}, considerando dos enfoques independientes. Por un lado, se utilizó \textit{Support Vector Machines} (SVM) sobre un conjunto de atributos definidos manualmente, incluyendo información proveniente de los \textit{embeddings} y de un léxico subjetivo. Por otro lado, se utilizó \textit{Convolutional Neural Networks} (CNN) consumiendo como entrada los \textit{embeddings} del \textit{tweet}. Finalmente se utilizó un enfoque híbrido basado en la combinación de los dos métodos anteriores, fundado en sus cualidades probabilísticas.

Los resultados muestran que los dos enfoques, SVM y CNN, dieron resultados interesantes en los diferentes corpus de evaluación, y que la combinación de los métodos aportó mejoras en casi todos los casos.

\section{Preprocesamiento}

La organización del TASS generó dos corpus de tweets anotados con su polaridad (\texttt{P}, \texttt{N}, \texttt{NEU} o \texttt{NONE}) para utilizar durante el entrenamiento: InterTASS y General TASS, de 1008 y 7219 tweets respectivamente. Más adelante se proveyó un nuevo corpus de desarrollo con características similares a InterTASS de 506 tweets. En nuestro trabajo decidimos utilizar ambos corpus combinados, y realizamos nuestra propia partición de los corpus en entrenamiento y desarrollo (validación), intentando mantener las mismas proporciones de tweets de cada categoría. Se separó el corpus General TASS en dos particiones: 85\% para entrenamiento y 15\% para desarrollo, y se unieron esos datos con las particiones de InterTass. El tamaño total de los corpus y las cantidades por categoría pueden verse en la tabla \ref{tab:tamano_corpus}.

\begin{table}[htbp]
    \centering
    \begin{tabular}{ l c c }
         & Entrenamiento & Desarrollo \\
        \midrule
        P    & 2782 (39\%) & 576 (36\%) \\
        N    & 2295 (32\%) & 524 (33\%) \\
        NEU  &  721 (10\%) & 151 (10\%) \\
        NONE & 1346 (19\%) & 338 (21\%) \\
        \midrule
        Total & 7144 & 1589
    \end{tabular}
    \caption{Tamaño de los corpus utilizados}
\label{tab:tamano_corpus}
\end{table}

Cada corpus fue preprocesado de la siguiente manera:

\begin{itemize}
    \item Se eliminan caracteres de espacio redundantes, referencias a URLs y elipsis.
    \item Todas las referencias a usuarios de Twitter se sustituyen por el token ``@user''.
    \item Se buscan repeticiones de un mismo carácter que aparezcan tres o más veces seguidas y se las sustituye por una sola instancia del carácter. Por ejemplo, sustituir ``holaaaa'' por ``hola''.
    \item Se sustituyen todas las interjecciones que denotan risa (por ejemplo ``jajajaja'', ``jejeje'', ``jajaj'') por el token ``jaja''.
    \item Se pasa todo el texto a minúsculas.
\end{itemize}

No se consideró información gramatical, como lema, categoría léxica, información morfológica o información sintáctica.

\section{Recursos utilizados}

\subsection{Léxicos de palabras positivas y negativas}

Se tomaron como punto de partida tres léxicos subjetivos disponibles para el español \cite{cruz2014building,saralegi2013elhuyar,brooke2009cross}, a partir de los cuales de generó un nuevo léxico, constituido por los elementos de la intersección de los tres léxicos originales, con un total de 312 lemas negativos y 301 lemas positivos. Este léxico fue expandido con las formas flexivas de cada lema, alcanzando un total de 4730 palabras. Esto permite paliar la decisión de no lematizar los tweets para su procesamiento. Para la expansión del léxico se utilizó el diccionario de FreeLing~\cite{padro12}.

\subsection{Colecciones de \textit{word embeddings}}

Se utilizaron los \textit{word embeddings} entrenados por \cite{AzzinnariTesis2016}. Fueron obtenidos mediante la técnica \emph{word2vec} \cite{mikolov2013efficient} y tienen dimensión 300. Están basados en un conjunto de datos de casi seis mil millones de palabras. Su procedencia es en mayoría texto de prensa extraído de Internet.

También se utilizaron \textit{word embeddings} entrenados a partir de la Wikipedia utilizando la técnica GloVe \cite{Pennington2014}, presentadas en \cite{Etcheverry2016}. Esta colección de vectores tuvo menor desempeño en nuestros experimentos, lo cual probablemente se deba a una menor cobertura léxica.

Finalmente se experimentó con el uso de \emph{fastText} \cite{bojanowski2016enriching} entrenándolo con la Wikipedia en español junto con 400,000 tweets en español provenientes de la API de Streaming de Twitter \emph{sample} para construir los \textit{embeddings}. Estos vectores fueron posteriormente descartados ya que no se obtuvo una mejora en los resultados para las pruebas realizadas.

\subsection{Predictor de polaridad por palabra}

    Se construyó un algoritmo de regresión basado en SVM utilizando los léxicos de palabras positivas y negativas como conjunto de entrenamiento. El objetivo es que este modelo sea capaz de obtener un valor real que represente la polaridad para cualquier palabra del lenguaje. El modelo toma como entrada los 300 valores reales del vector que representa la palabra y devuelve un valor real. Para el entrenamiento se consideró que las palabras positivas debían tener un valor positivo (1), y las negativas un valor negativo (-1).

\subsection{Marcadores de categoría}

    Se obtuvo una lista de todas las palabras del corpus de entrenamiento y para cada palabra se calculó la distribución de las categorías de todos los tweets en que aparece. Consideramos que una palabra es un \textit{marcador de categoría} si la proporción de veces que dicha palabra aparece en la categoría es mayor o igual a 75\%. Utilizando esta información, se confeccionaron listas de marcadores para las cuatro categorías: 429 positivas, 438 negativas, 12 neutras, y 33 marcadores de no opinión.

\section{Clasificadores}

\subsection{Enfoque basado en SVM}

Para la tarea 1, que consiste en evaluar la polaridad de un tweet, se construyó un clasificador SVM (Support Vector Machine) entrenado con los siguientes atributos:

\begin{itemize}
    \item Centroide de los \textit{embeddings} de las palabras del tweet. Trabajos anteriores han mostrado que el uso del centroide, o promedio de los \textit{embeddings}, si bien es una técnica muy simple, para algunos problemas da buenos resultados, en particular para el análisis de sentimiento~\cite{white2015well}. (300 valores reales)

    \item Polaridad de las nueve palabras principales del tweet según el clasificador de polaridad. La cantidad de palabras es el largo promedio de los tweets del corpus de entrenamiento, filtrando \textit{stopwords}. Se toma como más relevantes los valores de mayor valor absoluto.  Si el tweet tiene menos de nueve palabras se completan los nueve valores repitiendo las polaridades de las palabras del mismo tweet. (9 valores reales)

    \item Cantidad de palabras que pertenecen a los léxicos positivo y negativo. (2 valores naturales)

    \item Cantidad de palabras cuyas representaciones vectoriales están cercanas a los vectores promedio del léxico positivo y el léxico negativo. (2 valores naturales)

    \item Cantidad de palabras que pertenecen a las listas de marcadores de categoría. (4 valores naturales)

    \item Atributos que indican si el tweet original contiene repetición de caracteres o alguna palabra escrita completamente en mayúsculas. (2 valores booleanos)

    \item Cálculo de polaridad tentativa (P, N, NEU, NONE) en base a cantidad de palabras positivas y negativas, teniendo en cuenta el alcance de las negaciones. Se considera una lista de negadores que invierten la polaridad de las palabras, su alcance va desde el negador hasta algún signo de puntuación. (4 categorías)

    \item Cinco palabras más relevantes en el corpus de entrenamiento según \textit{bag of words}. El valor cinco se definió de manera heurística, filtrando las palabras pertenecientes a una lista de \textit{stopwords}, adaptada a la tarea de análisis de sentimientos (se eliminó de la lista de \textit{stopwords} un conjunto de palabras consideradas relevantes para el cálculo de la polaridad como ``no'', ``pero'', ``aunque'', etc.). Es interesante destacar que, en los experimentos realizados sobre el corpus de entrenamiento que utilizamos durante el desarrollo del trabajo, la lista de cinco palabras más relevantes obtenida fue: ``feliz'', ``gracias'', ``no'', ``pero'', ``portada'' (5 valores booleanos).
\end{itemize}

Este conjunto final de atributos utilizado para el entrenamiento del modelo se obtuvo a partir de varios experimentos, evaluando su desempeño sobre el corpus de desarrollo. Se comenzó utilizando solamente atributos de tipo \textit{bag of words}, comprobándose que el nivel de acierto y M-$F_1$ aumentaba utilizando una mayor cantidad de palabras, estancándose al llegar al entorno de las mil palabras más relevantes. Luego se entrenó un clasificador que utiliza el centroide de los \textit{embeddings}, obteniendo un mejor desempeño. Pero se comprobó que combinar ambos tipos de atributos (centroide y \textit{bag of words}) permitía obtener aún mejores resultados, aumentando el acierto hasta en 13 puntos y la M-$F_1$ hasta en 12 puntos. La combinación óptima encontrada utiliza los atributos del centroide más las cinco palabras más relevantes según \textit{bag of words}.

Los demás atributos considerados fueron aportando pequeños aumentos en los resultados a medida que se fueron incorporando. Si bien ninguno de estos atributos en particular aporta una mejora sustancial, la inclusión de todos ellos permite una aumento aproximado de 2 puntos en acierto y M-$F_1$, respecto a los resultados que se obtienen utilizando solamente el centroide y las cinco palabras más relevantes. En particular, se observa que el aporte del léxico subjetivo es poco relevante.

Para todos los experimentos con el modelo SVM se utilizó la librería \textit{scikit-learn} \cite{scikit-learn} de Python. Los modelos SVM de scikit-learn se entrenan de manera diferente dependiendo de si se habilita o no la salida con probabilidades por categoría. El cálculo de probabilidades para una salida multiclase se realiza utilizando el método de \cite{wu2004probability}, el cual es un proceso más lento y puede dar salidas diferentes a las obtenidas entrenando el modelo sin probabilidades. Para nuestros datos, en particular, se constató una mejora importante (2.5\%) en los valores de M-$F_1$ utilizando el entrenamiento con probabilidades, por lo que optamos por utilizar este modelo.

La tarea 2 consiste en clasificar la polaridad de tweets segmentada por aspectos. Para esta tarea se entrenaron dos clasificadores tipo SVM utilizando atributos similares a los utilizados para la tarea 1. El clasificador \texttt{svm1} utiliza atributos tipo \textit{bag of words} para los tweets, si el tweet original contenía repetición de caracteres o palabras en completamente en mayúsculas, cantidades de palabras que pertenecen a las listas de marcadores de categoría, y el aspecto sobre el que se está clasificando. El clasificador \texttt{svm2} utiliza solamente los 300 atributos reales que corresponden al centroide de los \textit{word embeddings} del texto del tweet y el aspecto sobre el que se está clasificando.

\subsection{Enfoque basado en CNN}

Se realizaron experimentos con redes neuronales para la tarea 1 tomando como entrada representaciones distribuidas de las palabras.

Con el objetivo de obtener una representación de la oración a partir de las representaciones vectoriales de las palabras, se consideraron dos enfoques. Por un lado, se consideró como línea base la concatenación de una ventana de palabras, enfoque similar a \namecite{Etcheverry2017}, aunque en una problemática distinta. Por otro lado se consideró la representación provista por convoluciones unidimensionales de las palabras del tweet,
basados en el trabajo de \namecite{Kim2014}, presentando este último resultados ampliamente superiores.

\begin{figure}[htbp!]
\centering
\includegraphics[width=0.4\textwidth]{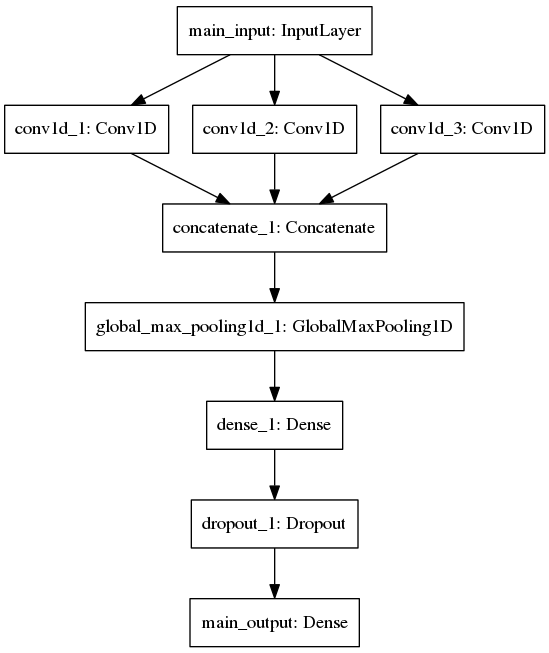}
\caption{
Arquitectura de red con varias ramas convolucionales.
}
\label{fig:conv_mb}
\end{figure}

Se consideraron dos variantes estructurales de modelos de redes neuronales convolucionales (CNN). Por un lado, modelos con una única capa convolucional y, por otro, modelos con varias capas convolucionales que reciben la entrada de forma independiente y sus salidas son concatenadas (ver fig. \ref{fig:conv_mb}). En ambos casos, luego de la representación provista por las convoluciones (o de la concatenación en el caso de varias ramas), se utilizó \textit{max pooling}\footnote{ Alternativamente, \textit{average pooling} fue considerado sin obtener mejores resultados.}, capas completamente conectadas, \textit{dropout} y \textit{softmax} en la salida.

Se utilizó la misma partición de entrenamiento y evaluación
mencionada anteriormente, separando $783$ tweets del corpus de entrenamiento para validación.
Todos los modelos fueron entrenados con \textit{adam} \cite{Kingma2014}, considerando un \textit{learning rate} de
$1\times10^{-4}$, $beta_1=9\times10^{-1}$, $beta_2=0.999$, $epsilon=1\times10^{-8}$ y $decay=0$.
Para evitar el sobreajuste de los modelos se utilizó \textit{dropout} (ver fig. \ref{fig:conv_mb}) y \textit{early stopping}.

Para la codificación se utilizó \textit{Keras} \cite{chollet2015keras}.
La codificación del modelo de una rama fue inspirada en el código provisto por François Chollet\footnote{\url{https://github.com/fchollet/keras/blob/master/examples/imdb_cnn.py}} y el de múltiples ramas en el de Chang-Uk Shin\footnote{\url{https://github.com/tflearn/tflearn/blob/master/examples/nlp/cnn_sentence_classification.py}}.

En las variantes estructurales consideradas para los modelos convolucionales, se observó mayor poder de generalización en la variante de múltiples ramas, por este motivo fue la utilizada para los resultados reportados (\texttt{cnn1}, ..., \texttt{cnn4}). El modelo final (denominado \texttt{cnn4}) consiste en 3 ramas convolucionales, con 2, 3 y 4 palabras en la convolución, 56 filtros y desplazamiento de a una palabra; una capa completamente conectada de dimensión 200; $0.2$ de probabilidad de pérdida para \textit{dropout}; y función de activación \textit{selu} \cite{Klambauer2017}.

\subsection{Enfoque híbrido}

Dado que tanto en el enfoque de SVM como en el de CNN entrenamos modelos que devuelven una distribución de probabilidades sobre las categorías predichas, utilizamos esta característica para entrenar un nuevo clasificador que combina las salidas de los dos clasificadores anteriores para la tarea 1. El nuevo clasificador devuelve la categoría $C$ para el tweet $T$ que maximice el promedio de las probabilidades de los clasificadores anteriores, como se muestra en la fórmula~\ref{eq:svm_cnn}.

\begin{equation}
    \argmax_{C}\frac{P_{SVM}(C|T) + P_{CNN}(C|T)}{2}
\label{eq:svm_cnn}
\end{equation}

\section{Análisis de resultados}

Los resultados para los experimentos realizados sobre el corpus de validación para la tarea 1 se muestran en la tabla~\ref{tab:resultados_val}. Como se puede observar, el modelo híbrido fue el que obtuvo mejores resultados. Se muestra la matriz de confusión para este modelo en la tabla~\ref{tab:matriz_confusion}.

\begin{table}[htbp]
    \centering
    \begin{tabular}{ c r r }
        Clasif. & M-$F_1$ & Acierto \\
        \midrule
        svm      & 49.9 & 61.7 \\
        cnn4      & 50.2 & 64.1 \\
        svm\_cnn & \textbf{50.9} & \textbf{64.7}
    \end{tabular}
    \caption{Resultados sobre el corpus de validación}
\label{tab:resultados_val}
\end{table}

\begin{table}[htbp]
    \centering
    \begin{tabular}{ r r r r r }
         & P & N & NEU & NONE \\
        \midrule
        P & 443 & 88 & 8 & 37 \\
        N & 85 & 414 & 5 & 20 \\
        NEU & 47 & 89 & 4 & 11 \\
        NONE & 89 & 78 & 4 & 167
    \end{tabular}
    \caption{Matriz de confusión para \texttt{svm\_cnn}. Las filas representan las clases reales, mientras que las columnas representan las clases predichas.}
\label{tab:matriz_confusion}
\end{table}

El modelo muestra buenos resultados para las clases \texttt{P}, \texttt{N} y \texttt{NONE}, pero no alcanza resultados aceptables para la clase \texttt{NEU}. Los tres modelos, y en particular este, fallan al intentar predecir los tweets neutros. Esto se puede deber a que hay pocas instancias pertenecientes a esta clase, siendo difícil poder aprender a partir de ellas. Otra particularidad que presentan los tweets neutros es el hecho de contener tanto elementos positivos como negativos, esto explicaría la tendencia del clasificador a confundirlos con esas dos clases, como muestra la matriz. Notar que para lograr mejorar los resultados de M-$F_1$ es fundamental mejorar la medida $F_1$ en los tweets neutros, ya que es la clase que da el número más bajo.

Los resultados para los experimentos realizados sobre los corpus de evaluación de la competencia se muestran en la tabla~\ref{tab:resultados_test}. A diferencia de lo constatado sobre el corpus de validación, el clasificador CNN obtuvo un desempeño menor que el SVM sobre los corpus de evaluación, lo cual puede indicar que la red se sobreajustó al corpus utilizado para entrenar. Por otra parte, en casi todas las evaluaciones la combinación entre SVM y CNN se desempeñó mejor que cada uno de los clasificadores independientes en los tres corpus.

\begin{table}[htbp]
    \centering
    \begin{tabular}{ l c r r }
        Corpus & Clasif. & M-$F_1$ & Acierto \\
        \midrule
        InterTASS & svm      & 45.7 & 58.3 \\
        InterTASS & cnn4      & 43.7 & 57.9 \\
        InterTASS & svm\_cnn & \textbf{47.1} & \textbf{59.6} \\
        \midrule
        Gral.\ TASS & svm      & 53.3 & 65.6 \\
        Gral.\ TASS & cnn4      & 53.1 & 66.5 \\
        Gral.\ TASS & svm\_cnn & \textbf{54.6} & \textbf{67.4} \\
        \midrule
        Gral.\ TASS\ 1k & svm      & \textbf{56.2} & 70.0 \\
        Gral.\ TASS\ 1k & cnn4      & 55.7 & 69.4 \\
        Gral.\ TASS\ 1k & svm\_cnn & 53.9 & \textbf{71.1}
    \end{tabular}
    \caption{Resultados sobre los corpus de evaluación}
\label{tab:resultados_test}
\end{table}

Se observa que los resultados sobre el corpus InterTASS son notoriamente más bajos que los obtenidos sobre los otros corpus. Esto puede deberse a que entrenamos nuestros sistemas con un conjunto único de tweets, provenientes en su mayoría del corpus General TASS, que contiene tweets con características diferentes a los contenidos en el InterTASS.

Los resultados de los clasificadores \texttt{svm1} y \texttt{svm2} para la tarea 2 se muestran en la tabla \ref{tab:resultados_task2}. El clasificador \texttt{svm2}, basado principalmente en el centroide de los tweets, es el que obtiene mejor desempeño en los dos corpus de evaluación.

\begin{table}[htbp]
    \centering
    \begin{tabular}{ l c r r }
        Corpus   &   Clasif.    &   M-$F_1$   &   Acierto \\
        \midrule
        social-TV   &   svm1    &   41.3   &   49.3 \\
        social-TV   &   svm2    &   \textbf{42.6}   &   \textbf{59.5} \\
        stompol     &   svm1    &   37.7   &   51.4 \\
        stompol     &   svm2    &   \textbf{50.8}   &   \textbf{59.0} \\
    \end{tabular}
    \caption{Resultados para la tarea 2}
\label{tab:resultados_task2}
\end{table}

\section{Conclusiones}

Este trabajo muestra tres enfoques para la tarea 1 del TASS 2017 sobre análisis de sentimiento de tweets en español: SVM con atributos elaborados a mano, que incluyen información proveniente de los vectores de palabras; redes neuronales con un fuerte uso de vectores de palabras; y un enfoque híbrido, que combina los dos enfoques anteriores. El enfoque que funcionó mejor fue el híbrido. En cuanto a la tarea 2, que trata sobre la clasificación de la polaridad de los tweets segmentados por aspectos, el enfoque SVM utilizando \textit{embeddings} mostró tener buenos resultados, aunque aún queda mucho espacio de mejora.

En base al análisis experimental se observa que el sistema es poco efectivo en la detección de los tweets clasificados como neutros. El hecho de que esta clase tenga pocos representantes en los corpus y su similitud tanto con los tweets negativos como con los positivos podrían ser las causas del bajo desempeño en su detección.

\bibliographystyle{fullname_esp}
\bibliography{refs}

\end{document}